\documentclass[conference]{IEEEtran}
\usepackage{times}

\usepackage[numbers]{natbib}
\usepackage{multicol}
\usepackage[bookmarks=true]{hyperref}
\usepackage{amsfonts} 
\usepackage{graphicx}
\usepackage{amsmath}
\usepackage{booktabs} 
\usepackage{multirow}
\usepackage{amssymb}
\usepackage{algorithm}
\usepackage{algorithmic}
\usepackage{xurl}
\usepackage[table]{xcolor}
\usepackage{caption}
\usepackage{multirow}
\usepackage{fontawesome5}
\usepackage{textcomp} 
\newcommand{\tildemid}{\raisebox{0.5ex}{\texttildelow}}

\definecolor{green}{RGB}{11,155,13}

\begin{document}

\title{\LARGE \texttt{Verti-Bench}: A General and Scalable Off-Road Mobility Benchmark for Vertically Challenging Terrain}

\author{Tong Xu, Chenhui Pan, Madhan B. Rao, Aniket Datar, Anuj Pokhrel, Yuanjie Lu, and Xuesu Xiao\\
George Mason University}



%

\makeatletter
\g@addto@macro\@maketitle{
  \begin{figure}[H]
  \setlength{\linewidth}{\textwidth}
  \setlength{\hsize}{\textwidth}
  \centering
  \includegraphics[width=\textwidth]{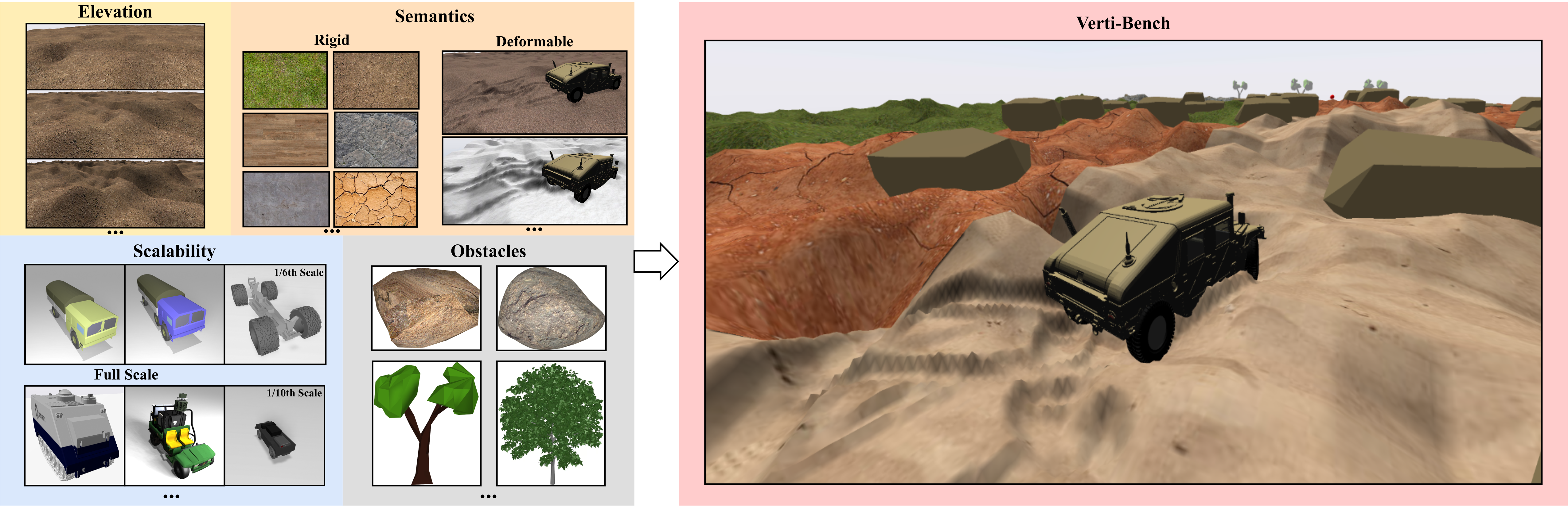}
  \caption{Based on high-fidelity multi-physics simulation, Verti-Bench encapsulates a variety of off-road features, i.e., geometry, semantics, and obstacles, and vehicles and can be scaled to different sizes. 100 off-road environments and 1000 navigation tasks with millions of off-road features can objectively and quantitatively evaluate off-road vehicle mobility on extremely rugged, vertically challenging terrain.}
    \label{fig:cover}
  \end{figure}
}
\makeatother

\maketitle

\addtocounter{figure}{-1}

\begin{abstract}
Recent advancement in off-road autonomy has shown promises in deploying autonomous mobile robots in outdoor off-road environments. 
Encouraging results have been reported from both simulated and real-world experiments. 
However, unlike evaluating off-road perception tasks on static datasets, benchmarking off-road mobility still faces significant challenges due to a variety of factors, including variations in vehicle platforms and terrain properties. 
Furthermore, different vehicle-terrain interactions need to be unfolded during mobility evaluation, which requires the mobility systems to interact with the environments instead of comparing against a pre-collected dataset. 
In this paper, we present Verti-Bench, a mobility benchmark that focuses on extremely rugged, vertically challenging off-road environments. 
100 unique off-road environments and 1000 distinct navigation tasks with millions of off-road terrain properties, including a variety of geometry and semantics, rigid and deformable surfaces, and large natural obstacles, provide standardized and objective evaluation in high-fidelity multi-physics simulation. 
Verti-Bench is also scalable to various vehicle platforms with different scales and actuation mechanisms. 
We also provide datasets from expert demonstration, random exploration, failure cases (rolling over and getting stuck), as well as a gym-like interface for reinforcement learning. 
We use Verti-Bench to benchmark ten off-road mobility systems, present our findings, and identify future off-road mobility research directions. Verti-Bench project website can be found at \textcolor{magenta}{\href{https://cs.gmu.edu/~xiao/Research/Verti-Bench}{\texttt{https://cs.gmu.edu/\tildemid{}xiao/Research/Verti-Bench}}}.

\end{abstract}
\IEEEpeerreviewmaketitle

\section{Introduction}
\label{sec::intro}

Off-road autonomous mobile robots present unique opportunities in search and rescue, environment monitoring, scientific exploration, and other application domains.  
However, off-road autonomy presents unique challenges to both robot perception and vehicle mobility that distinguish from structured on-road environments~\cite{jackel2006darpa, naranjo2016autonomous, price2024expanding}, such as variable geometry, deformable surfaces, and natural obstacles. Considering recent advancement, standardized benchmarks for off-road autonomy are necessary to objectively and quantitatively evaluate and compare the progress of the off-road robotics community. 

Unlike off-road perception, evaluating off-road mobility is more difficult. A plethora of off-road perception datasets~\cite{wigness2019rugd, jiang2021rellis, triest2022tartandrive, min2022orfd, mortimer2024goose, knights2023wild, sharma2022cat, liu2024botanicgarden} are available to provide static ground truth labels to evaluate against perception systems' outputs, since the robot actions can be recorded and fed into the perception systems, not generated by them  (if actions are necessary at all). However, mobility systems produce new actions to drive robots to different states than those collected in the dataset, i.e., distribution shift, and thus cannot be evaluated by comparing against a static dataset. Therefore, vehicle-terrain interactions need to be unfolded during off-road mobility evaluation, which creates difficulty in standardization across research groups. 

Considering a lack of standard off-road mobility evaluation, researchers currently develop their own benchmarks to evaluate their mobility systems and re-implement previous systems to compare against. Such a practice, however, leads to ad-hoc evaluation in a few aspects: The evaluation platforms, either a physical or a simulated robot, vary in terms of robot size, weight, actuation mechanism, and/or levels of off-road physics simulation fidelity; The evaluation environments range from handcrafted off-road simulation features~\cite{rana2024towards, yu2024adaptive, young2020unreal, xu2024reinforcement, cai2025pietra, so2022sim}, small-scale indoor testbeds~\cite{xu2023efficient, datar2024toward, xiao2018review, xiao2015locomotive}, enclosed outdoor tracks~\cite{goldfain2019autorally, xiao2021learning, atreya2022high, karnan2022vi, pan2020imitation}, and large-scale real-world testing facilities~\cite{triest2022tartandrive, han2023model, jiang2021rellis}; Re-implementation of previous approaches on one's own robot is not only laborious, but also subject to misinterpretation of implementation details. Therefore, a standard off-road mobility benchmark which is general and scalable to all these aspects is desired for the off-road mobility research community. 

In this work, we present Verti-Bench (Fig.~\ref{fig:cover}), a general and scalable off-road mobility benchmark that focuses on extremely rugged, vertically challenging terrain with a variety of unstructured off-road features. The main goal of Verti-Bench is to directly compare the performances of different off-road mobility systems. Based on a high-fidelity multi-physics dynamics simulator, Chrono~\cite{tasora2016chrono}, Verti-Bench encapsulates variations in four orthogonal dimensions: Using the Sliced Wasserstein Autoencoder (SWAE)~\cite{kolouri2018sliced} and real-world off-road terrain data, off-road geometry is represented as a diverse set of 2.5D elevation maps; Ten terrain semantics classes, including seven rigid and three deformable, are designed with different distributions of physics parameters, e.g., friction coefficient, cohesive effect, and soil stiffness; Different types of natural obstacles, e.g., boulders and vegetations, are randomly distributed based on different densities; A set of off-road vehicles with different scales (from 1/10th to full scale) and actuation mechanisms (4-, 6-, and 8-wheeled and 2-tracked chassis, single- and double-wishbone, multilink, toebar leaf-spring, and special tensioning suspensions, as well as pitman-arm, rack-and-pinion, toebar, bellcrank/rotary arm, and differential steering) are provided, with the possibility of adding customized vehicles. Using Verti-Bench, we also provide datasets from expert demonstration, random exploration, failure cases (rolling over and getting stuck), as well as a gym-like interface for Reinforcement Learning (RL). We use Verti-Bench to benchmark ten off-road mobility systems, present our findings, and identify future off-road mobility research directions. To summarize, our contributions are: 
\begin{itemize}
    \item A general off-road mobility benchmark on vertically challenging terrain with 100 off-road environments and 1000 navigation tasks scalable to various vehicle types;
    \item Millions of off-road terrain features including geometry, semantics (rigid and deformable), and obstacles;
    \item Various datasets and a RL interface to facilitate data-driven off-road mobility;
    \item Findings and future research directions based on benchmark results of various off-road mobility systems.  
\end{itemize}

\section{Related Work}
\label{sec::related}
In this section, we review the field of off-road autonomy and current evaluation practices to motivate Verti-Bench.

\subsection{Off-Road Autonomy}
Starting from the DARPA LAGR program~\cite{jackel2006darpa}, roboticists have started developing autonomous systems to operate in off-road enviornments. Compared to indoor or on-road operations, off-road environments pose significant challenges during the entire sense-plan-act loop: Robot state estimation systems, such as visual inertial odometry~\cite{forster2016manifold} and simultaneous localization and mapping~\cite{stachniss2016simultaneous}, are easily affected by the unstructured visual features from the off-road environments as well as noisy inertial signals caused by the extensive vehicle vibrations; Perception is more than a geometric problem, i.e., free vs. obstacle spaces, as in indoor or on-road settings, and requires the consideration of semantics, e.g., gravel vs. grass. Both terrain geometry and semantics need to be represented~\cite{miki2022elevation, ewen2022these, dashora2022hybrid, sikand2022visual} for downstream planning and control tasks; Off-road planners and controllers need to reason beyond collision avoidance and consider factors such as vehicle stability~\cite{pokhrel2024cahsor, bae2021curriculum}, wheel slippage~\cite{siva2019robot, sharma2023ramp, siva2022nauts}, and terrain traversability~\cite{castro2023does, cai2024evora, seo2023learning, fan2021step}, oftentimes in $\mathbb{SE}(3)$ instead of $\mathbb{SE}(2)$~\cite{datar2024toward} due to the uneven off-road terrain surfaces~\cite{lee2023learning, datar2024toward}. 
The recent increase in interest in off-road autonomy~\cite{xiao2022motion} necessitates standard benchmarks to objectively and quantitatively evaluate research progress from the entire community, which is the motivation behind Verti-Bench. 

\subsection{Evaluating Off-Road Perception}
Off-road perception research still dominates the majority of the body of off-road autonomy work~\cite{xiao2022motion}. Fortunately, evaluating off-road perception can be mostly carried out on pre-collected, static datasets in a vehicle-agnostic fashion. 
For both evaluation and training of perception systems in a data-driven manner, a variety of off-road perception datasets are available for research in semantics segmentation~\cite{wigness2019rugd, jiang2021rellis, mortimer2024goose}, freespace detection~\cite{min2022orfd}, place recognition~\cite{knights2023wild}, traversability estimation~\cite{sharma2022cat}, and map reconstruction~\cite{liu2024botanicgarden}. Given a new off-road perception system taking the recorded sensor data and robot actions as inputs, its outputs can be simply compared against the ground truth labels provided by those datasets. A performance metric, e.g., segmentation accuracy, detection rate, recognition count, traversability correctness, and reconstruction precision, can then be computed to quantify how well the new perception system compares against others. Unfortunately, off-road mobility evaluation cannot be conducted on such pre-collected, static datasets. Notice that while off-road dynamics datasets~\cite{triest2022tartandrive} can be used to learn off-road navigation models, they cannot be used to evaluate off-road mobility.

\subsection{Evaluating Off-Road Mobility}
To the best of our knowledge, no standard off-road mobility benchmarks currently exist in the literature. 
Unlike off-road perception evaluation, mobility evaluation requires the vehicle-terrain interactions to be unfolded, since a different action will lead the robot into a different state absent from the dataset. Such a distribution shift necessitates mobility evaluation to be based on the unfolded, not pre-collected, vehicle trajectories.

To unfold vehicle-terrain interactions for objective off-road mobility evaluation, the vehicle platforms and terrain properties are required to be standardized, which, unfortunately, vary significantly across different research groups: Robots of different sizes, weights, actuation mechanisms (e.g., wheeled, tracked, steered, and differential-driven) are used to compare new mobility systems with existing ones. The latter is usually customized to fit for a different robot, potentially causing misinterpretation of implementation details; Different simulators, e.g., Gazebo~\cite{rana2024towards}, IsaacGym~\cite{yu2024adaptive}, Unreal~\cite{young2020unreal}, and Unity~\cite{so2022sim}, are leveraged to train and evaluate off-road mobility systems. To improve simulation speed, most of those simulators do not focus on physics fidelity, which is crucial for off-road mobility on extremely rugged and deformable surfaces; In the real world, small-scale indoor testbeds have been set up with foam, rocks, plywood, and pipes to emulate vertically challenging terrain in the wild~\cite{xu2023efficient, datar2024toward, xiao2018review, xiao2015locomotive}; Enclosed outdoor off-road tracks have been constructed to evaluate aggressive autonomous driving~\cite{goldfain2019autorally, xiao2021learning, atreya2022high, karnan2022vi, pan2020imitation}; A few research groups have access to large-scale off-road testing facilities with full-size vehicles~\cite{triest2022tartandrive, han2023model, jiang2021rellis}. 

With such diverse setups, an objective evaluation and comparison across different off-road mobility systems become infeasible. Verti-Bench, based on a high-fidelity multi-physics dynamics simulator, is hence motivated to fill such a gap of missing standard off-road mobility benchmarks. Notice that Verti-Bench is not meant to replace existing evaluation setups, but to complement them with a new standardized option to facilitate fair comparison across off-road mobility systems.


\section{Verti-Bench}
\label{sec::approach}

We present Verti-Bench's core high-fidelity multi-physics dynamics engine, diverse set of off-road features, including wide-ranging geometry, physics-grounded semantics, and natural obstacles, scalability to a variety of vehicle platforms, and standardized metrics to quantify off-road mobility performance. We also discuss various datasets we collect using Verti-Bench to complement real-world off-road mobility data to develop data-driven systems.

\subsection{Simulation}
Verti-Bench is based on Project Chrono~\cite{tasora2016chrono}, a high-fidelity multi-physics dynamics engine with a platform-independent open-source design implemented in C++ with a Python version, PyChrono. Compared to other commonly used robotics simulators (e.g., Gazebo, Unreal, Unity, PyBullet, MuJoCo, and IsaacGym with well-known physics limitations especially for differential-drive mobile robots~\cite{zifan_isaac}), Chrono is especially suitable to simulate complex off-road vehicle-terrain interactions involving suspension, tire, track, and terrain deformation, varying terrain contact friction, vehicle weight distribution and momentum, motor, powertrain, transmission, and wheel torque characteristics, aggressive vehicle poses with all six Degrees of Freedom (DoFs), etc. 
In Chrono, vehicle systems and terrain properties are made of rigid and flexible/compliant parts with constraints, motors and contacts, along with three-dimensional shapes for collision detection. 

One point worth noting is that Verti-Bench's choice of Chrono as its core simulator is primarily due to its high-fidelity multi-physics dynamics, a vital aspect for off-road mobility evaluation. However, Chrono is not the best simulator for photorealism, one focus of off-road perception simulation, which is out of scope of Verti-Bench. For the perception components, Verti-Bench has standard interfaces to provide ground truth elevation and semantics maps and obstacle occupancy grids. Another point is that Chrono is not yet GPU-accelerated. Combined with the high computation required for high physics fidelity, Chrono can only provide slightly faster-than-real-time simulation, depending on the complexity of the simulated environments (e.g., areas of deformable terrain and number, size, and number of mesh vertices of obstacles). Therefore, despite its intended efficient usage in off-road mobility evaluation, learning off-road mobility is expected to take a significant amount of training time with Verti-Bench (e.g., using our provided gym-like RL interface). 

In Chrono, each of the 100 Verti-Bench full-scale environments is constructed as a 129m$\times$129m world, with a resolution of 1m per pixel. Each environment can be down-scaled to cater vehicles of different sizes, e.g., 1/6th or 1/10th scale. Each pixel contains geometry, semantics, and obstacle information (details below). For each of the 100 environments, ten pairs of start and goal locations separated by 120 m are distributed in a circular manner, leading to a total of 1000 navigation tasks. 

\subsection{Geometry}
Real-world off-road terrain is characterized by various geometry in terms of elevation changes, e.g., slopes, hills, ditches, gullies, ravines, and other form of undulations. Some of such terrain can be traversed by certain types of off-road vehicles, while others cannot. Autonomous off-road mobility systems need to decide which of them can be attempted with what vehicle maneuvers. For example, a steep slope with low friction cannot be traversed at low speeds, but large vehicle momentum by high speeds at the bottom can help the vehicle ascend the top; Approaching a deep ditch quickly may get the vehicle stuck due to extensive suspension depression at the bottom, but slowly negotiating through is possible to mitigate suspension travel in order to maximize clearance.  

\begin{figure}[ht]
    \centering
    \includegraphics[width=\columnwidth]{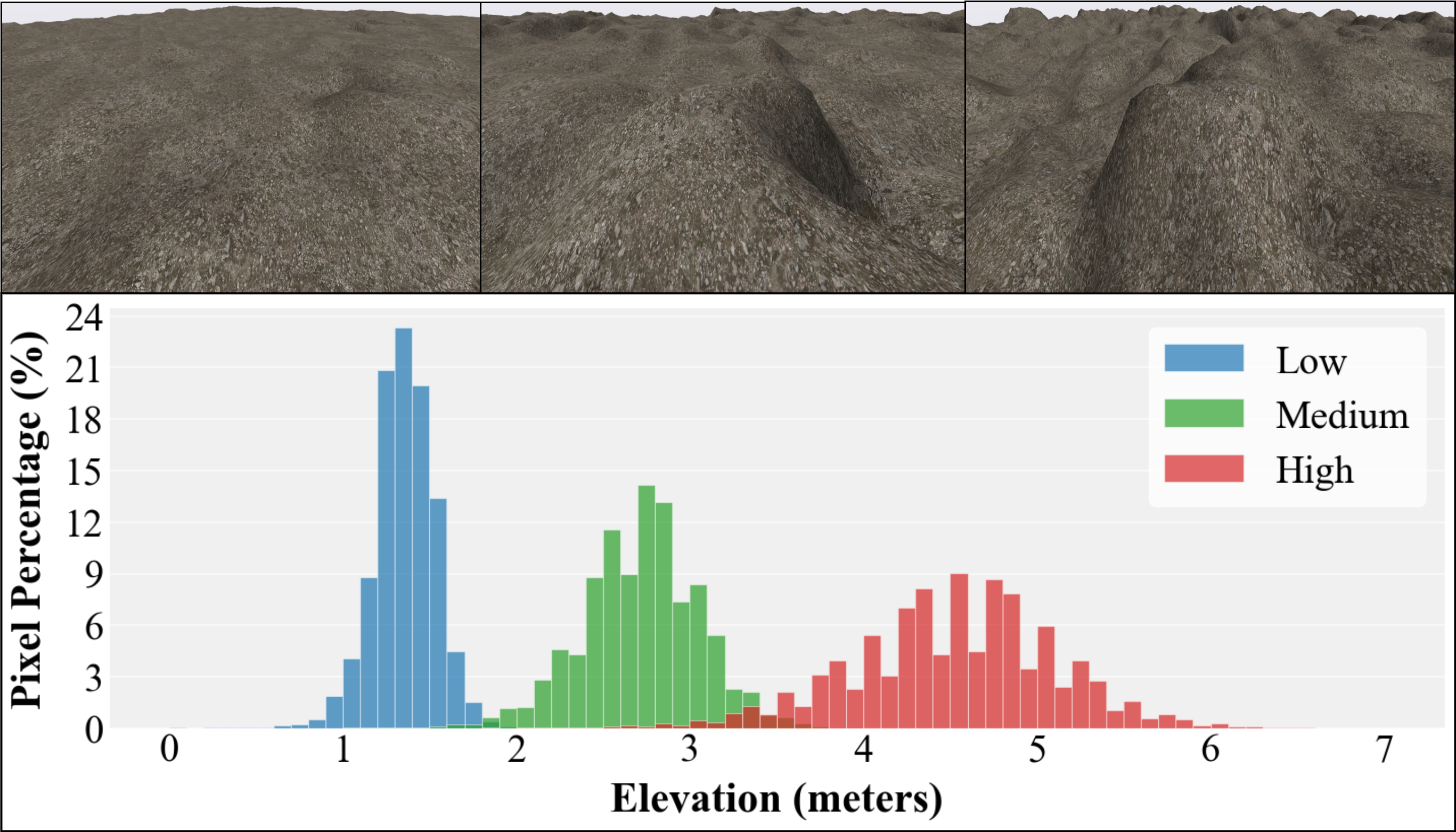}
    \caption{Top: Low, Medium, and High Elevation Maps; Bottom: Elevation Histograms across Three Elevation Levels. }
    \label{fig::elevation}
\end{figure}

Therefore, the geometry of Verti-Bench environments is represented as 2.5D elevation maps created by SWAE~\cite{kolouri2018sliced} and real-world elevation data. To be specific, we physically construct vertically challenging terrain with boulders and rocks and use a Microsoft Azure Kinect RGB-D camera to create elevation maps of different real-world terrain surfaces~\cite{mikielevation2022}. We then use SWAE~\cite{kolouri2018sliced}, a scalable generative model that captures the rich and often nonlinear distribution of high-dimensional data, as a feature extractor to reduce the dimension of the real-world elevation maps while preserving the original elevation information in a latent space, from which samples can be drawn to generate new elevation maps that resemble real-world vertically challenging terrain. To further introduce diversity and quantification of Verti-Bench geometry, we scale the output of the trained SWAE to 30\%, 60\%, and 100\% and denote them as low, medium, and high elevation level (Fig.~\ref{fig::elevation} top). Each Verti-Bench environment is generated with 1/3 probability of each elevation level. Fig.~\ref{fig::elevation} bottom shows the histogram of elevation values of all three levels of terrain geometry. High elevation environments also have the largest variance (most rugged terrain), while low elevation environments are smoother. 

\subsection{Semantics}
In addition to geometry, off-road terrain also presents challenges in terms of semantics and its associated physical vehicle-terrain contact features, such as friction, slip, and deformability. 
For example, an autonomous off-road mobility system should be aware that when driving through an icy or sandy laterally inclined slope with low friction or high deformability, sideway sliding downhill or wheel sinkage due to imbalanced load and then rollover is possible, respectively.  

\begin{figure}[h]
    \centering
    \includegraphics[width=\columnwidth]{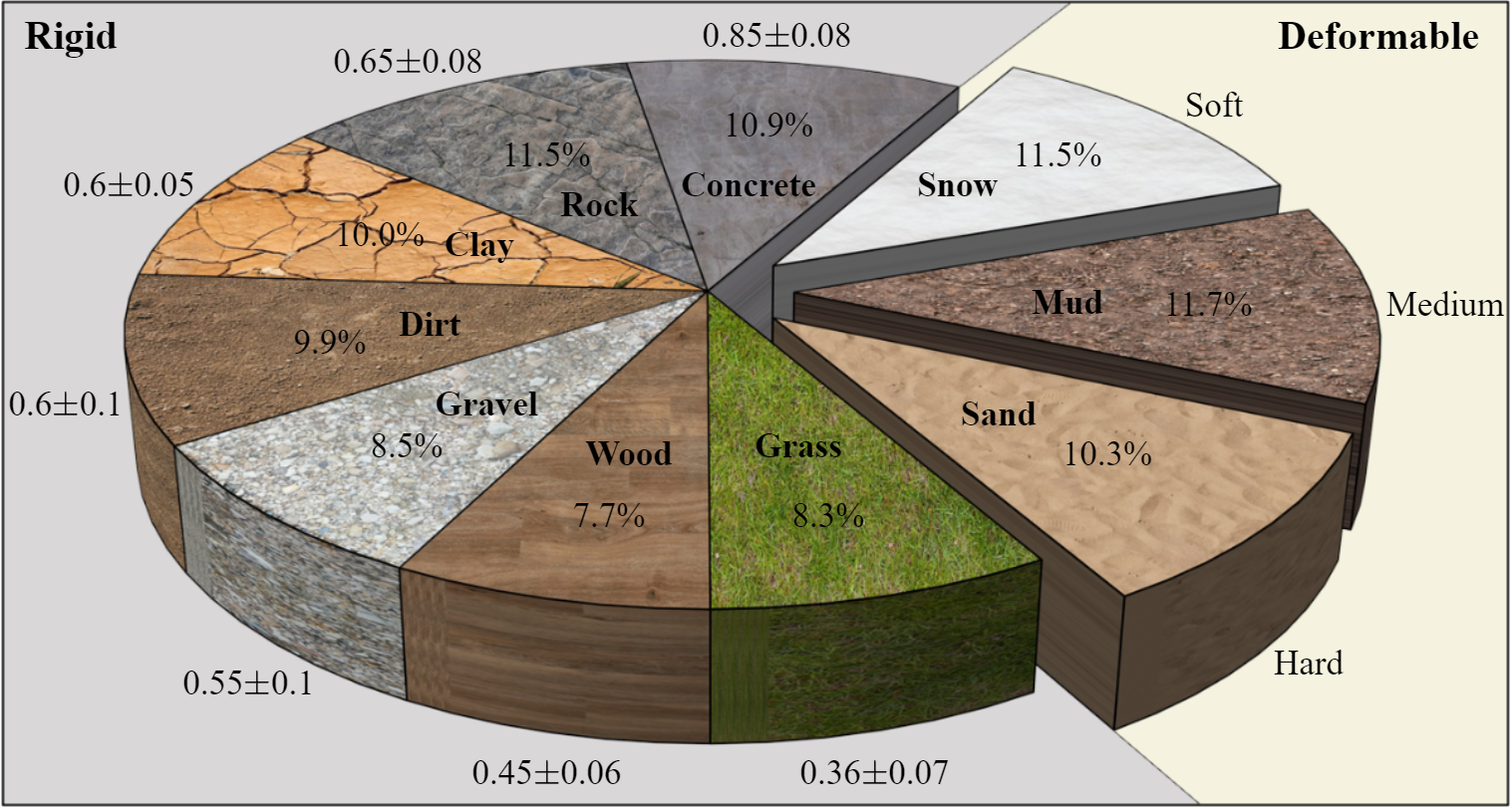}
    \caption{Seven Rigid (percentage, mean and variance of friction coefficient, and texture) and Three Deformable (percentage, deformability, and texture) Terrain Semantics.} 
    \label{fig::semantics}
\end{figure}

Therefore, we also add ten different semantics classes to the terrain elevation. We design seven rigid and three deformable semantics classes with different textures and distributions of physics parameters (Fig.~\ref{fig::semantics}). To be specific, the seven rigid semantics classes, i.e., grass, wood, gravel, dirt, clay, rock, and concrete, associate with a normal distribution of friction coefficient. When a pixel is sampled to be a certain terrain type, its friction coefficient is sampled from the corresponding distribution. We fix the restitution coefficient to 0.01 for all rigid semantics classes. For the three deformable terrain classes, i.e., snow, mud, and sand, we adopt the deformable Soil Contact Model (SCM) based on the Bekker-Wong model~\cite{laughery1990bekker} to simulate terrain deformation after wheel interaction: SCM presents the underlying terrain by a 2D grid and assumes each cell can only be displaced vertically and does not maintain any history other than the current vertical displacement. We hard-code three sets of physics parameters, including cohesive effect, soil stiffness, and hardening effect, for three different deformability levels, i.e., soft, medium, and hard. 
Verti-Bench also provides terrain with granular materials. But due to the slow simulation speed when simulating thousands of particles, it is only reserved for special evaluation circumstances where granular materials must be simulated and simulation speed is not of concern. All statistics of the ten terrain semantics classes can be found in Fig.~\ref{fig::semantics}. 

To create various terrain semantics while maintaining simulation efficiency, each 129$\times$129 Verti-Bench environment is first partitioned into a 16$\times$16 grid, with each grid cell as a 9$\times$9 patch (one overlapping pixel between every pair of patches to assure connectivity).  
To emulate real-world continuous terrain patches with same semantics and similar physical properties, we employ a cluster-based approach, where cluster centers are sampled from the environment. 
Each patch is then associated with its nearest cluster center using Euclidean distance. For all patches associated with a cluster center, the same semantics class is sampled and corresponding texture assigned, with each patch's physical property sampled from a predetermined distribution (Fig.~\ref{fig::semantics}). 
This approach creates natural physical variations within every region of the same semantics class while maintaining semantics diversity across regions.

\subsection{Obstacles}
\begin{figure}[ht]
    \centering
    \includegraphics[width=\columnwidth]{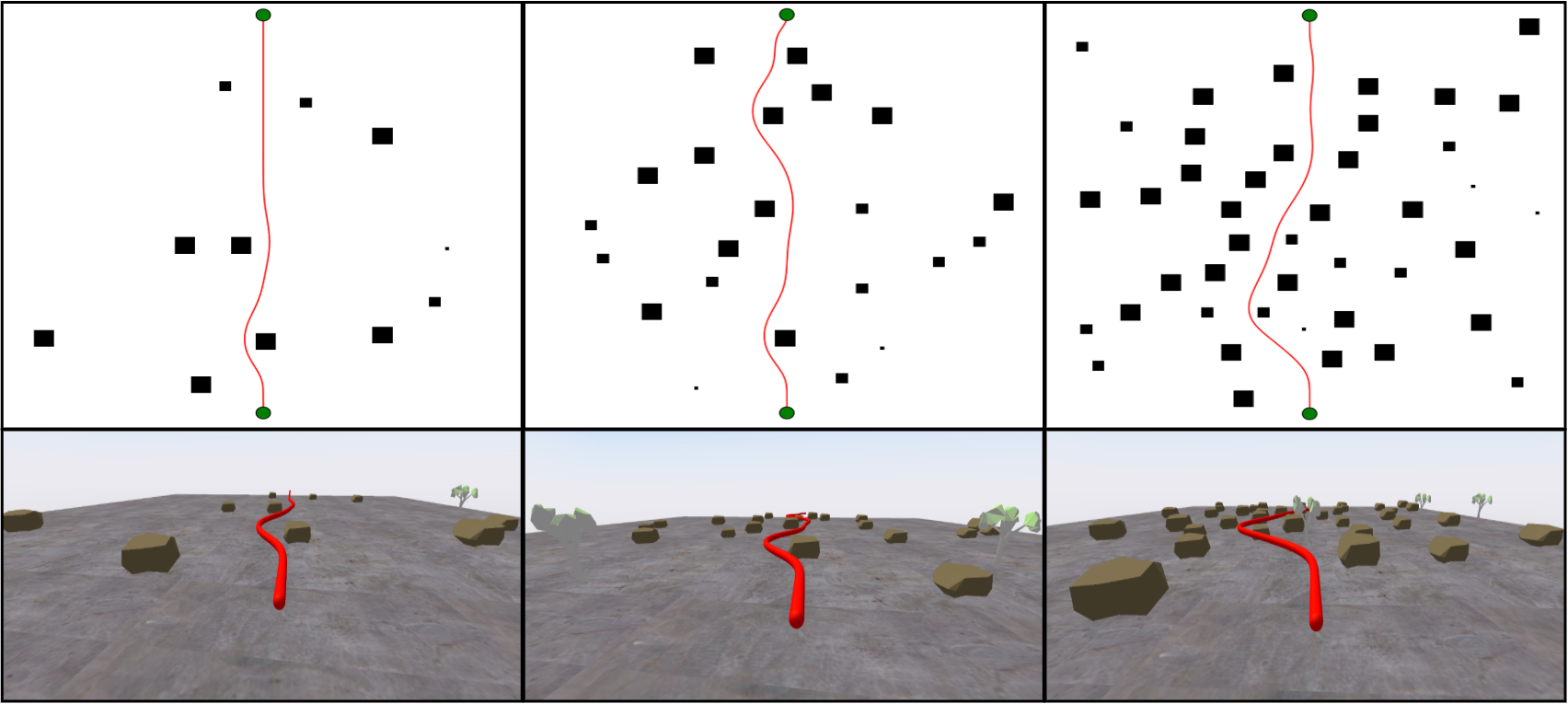}
    \caption{Top: Sparse, Medium, and Dense Obstacles (black) and Global Paths (red) between Start and Goal (green). Bottom: Corresponding simulation scenario in Verti-Bench (elevation and semantics are removed for obstacle clarity).}
    \label{fig::obstacles}
\end{figure}

Undulating geometry and varying semantics require off-road mobility systems to understand fine-grained vehicle-terrain interactions when driving on them. Off-road obstacles, like large boulders or trees, exist in real-world off-road environments, which are simply beyond vehicles' mechanical capabilities and hence need to be avoided. We also include natural obstacles in Verti-Bench to pose challenges to obstacle avoidance systems. For example, a large boulder triple the size of the vehicle is completely non-traversable, while a steep hill as part of the terrain may or may not be ascended with the right maneuver. We add natural obstacles as instances of the former. To further promote variation, we randomly sample the locations and types (different sizes of boulders or trees) of 10, 20, and 40 obstacles to place on each 129$\times$129 Verti-Bench environment, denoted as sparse, medium, and dense for obstacle distribution. We resample a new obstacle if the old one is within 10 m of another obstacle, a start, or a goal. Assuming a holonomic point-mass vehicle, we also provide pre-planned global paths leading from start to goal locations and avoiding obstacles. Fig.~\ref{fig::obstacles} shows three examples of sparse, medium, and dense obstacle distributions and their corresponding global paths. 

\subsection{Vehicles}
\begin{figure*}[ht]
    \centering
    \includegraphics[width=2\columnwidth]{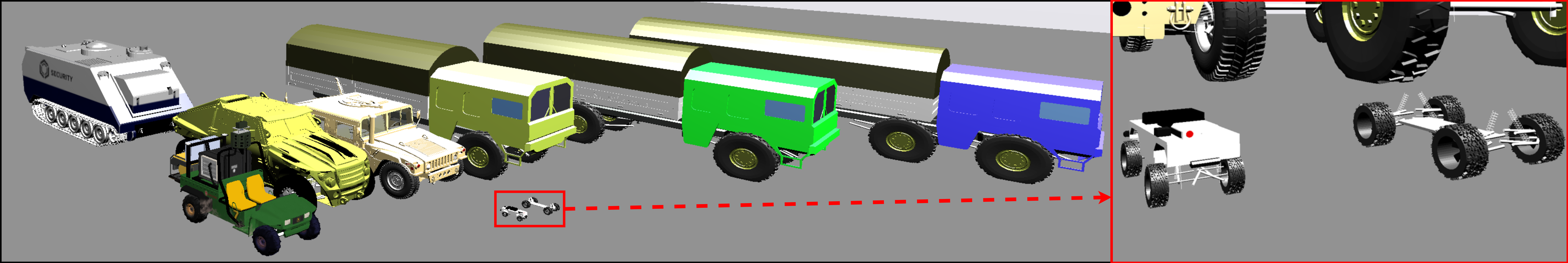}
    \caption{Verti-Bench Vehicles with Different Scale (1/10th, 1/6th, and full scale), Chassis (4-, 6-, and 8-wheeled and 2-tracked), Steering (pitman-arm, rack-and-pinion,
toebar, bellcrank/rotary arm, and differential), and Tires (rigid
and handling). }
    \label{fig::vehicles}
\end{figure*}

We also provide a set of vehicle platforms in Verti-Bench, with the possibility of adding new customized ones in the future, so that different off-road mobility systems can be evaluated on standardized vehicles. Compared to simplified vehicles in existing simulators, the Verti-Bench vehicles are more sophisticated and articulated, including engine/motor, drivetrain, transmission, suspension, steering mechanism, and wheel torque, whose responses to complex terrain interactions are simulated. To be specific, Verti-Bench  provides nine types of off-road vehicles, which are sourced from Project Chrono~\cite{tasora2016chrono}, open-source real and simulated research platforms~\cite{elmquist2022art}, and custom-created vehicles using 3D scanning and modeling (with a Creality CR-Scan Raptor 3D scanner) of real-world scaled vehicles (Fig.~\ref{fig::vehicles}). Those vehicles vary in terms of scale (1/10th, 1/6th, and full scale), chassis (4-, 6-, and 8-wheeled and 2-tracked), suspension (single- and double-wishbone, multilink, toebar leaf-spring, and special tensioning), steering (pitman-arm, rack-and-pinion, toebar, bellcrank/rotary arm,
and differential), and tires (rigid and handling, excluding FEA-based models due to significantly reduced simulation speed). All vehicles, regardless of their sources, are implemented as native C++ classes in Chrono's C++ framework. The C++ implementations are then compiled and exposed to Python through SWIG-generated bindings.  


\subsection{Metrics}
Verti-Bench automatically computes a set of standard metrics to quantify off-road mobility performance, while additional metrics can be customized as needed: A successful completion is defined as the robot reaching within 10 meters of the goal in less than 60 seconds. A successful trial may include contacts with obstacles. Success Rate captures the percentage of successful trials over total number of attempts; Traversal Time indicates how long it takes to finish a successful traversal; Roll and Pitch describes the stability of the vehicle during traversal, whose raw and absolute values can be used to derive mean, variance, and maximum; Vehicle actions, such as throttle and steering, can be used as metrics to quantify energy consumption, planning confidence, and path smoothness, along with other metrics. Verti-Bench provides infrastructure to save raw vehicle-terrain interaction data as well as to compute performance metrics. 

\subsection{Datasets}
Using Verti-Bench, we also collect a few datasets to facilitate future data-driven off-road mobility research. Current datasets are collected on the High Mobility Multipurpose Wheeled Vehicle (HMMWV, Fig.~\ref{fig:cover} right), while future data collection can expand to different vehicles. 

\subsubsection{Expert Demonstration} 
A team of four human operators collect 4 hours of expert demonstration of successfully driving the off-road vehicle in different Verti-Bench environments. We filter out all failure cases, e.g., vehicle rollover and getting-stuck, to maintain high demonstration quality. 

\subsubsection{Random Exploration}
To facilitate off-road kinodynamics learning~\cite{triest2022tartandrive}, we collect a random exploration dataset on different off-road terrain by driving the off-road vehicle with sinusoidal steering and 2 m/s speed commands for ten hours. Each data collection trial terminates if the vehicle rolls over, gets stuck, or reaches the environment boundaries.

\subsubsection{Failure Cases}
To enable future data-driven off-road mobility by preventing vehicle failures, we also curate a dataset of failure cases by providing the last ten seconds of trajectory before the vehicle rolls over or gets stuck. We divide all failure cases into two categories: (1) Rollover: the robot has a $>30$\textdegree~roll angle at the end of a failed trial; (2) Stuck: the robot does not move more than 1 m during the last 10 seconds of a failed trial (with $\leq30$\textdegree~roll). The failure cases can be used to learn high-cost regions to be avoided in data-driven mobility systems. 

\subsubsection{RL Interface}
Although not being a main purpose of Verti-Bench, we also provide a gym-like RL interface so that vehicles can learn off-road mobility through trial-and-error experiences in Verti-Bench. Vehicle states and actions, as well as reward functions, can be customized by our interface, which communicates with existing RL algorithm implementations. Verti-Bench provides interfaces of different inputs to the mobility systems, such as elevation and semantic maps, robot states, and raw sensor data.  Users can choose appropriate inputs based on their evaluation needs.
\section{Evaluation and Discussions}
\label{sec::evaluation}


\begin{figure*}[h!]
    \centering
    \includegraphics[width=0.49\textwidth]{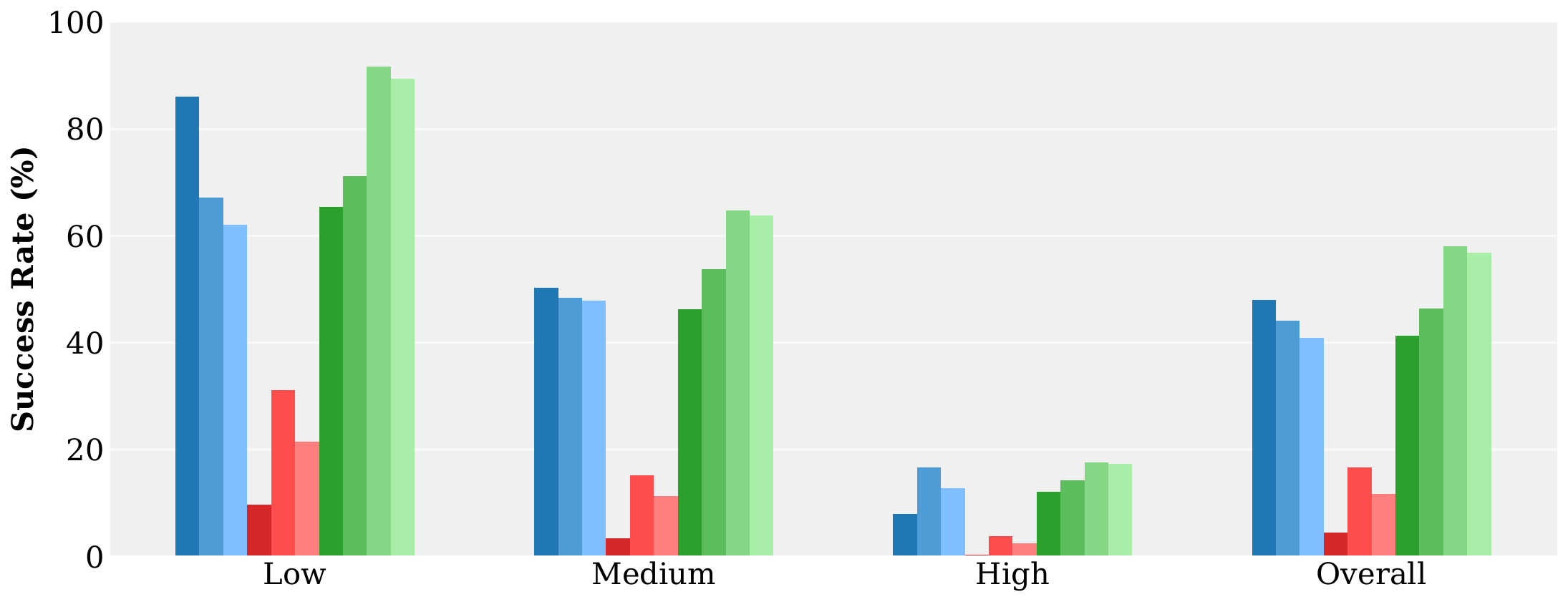}
    \includegraphics[width=0.49\textwidth]{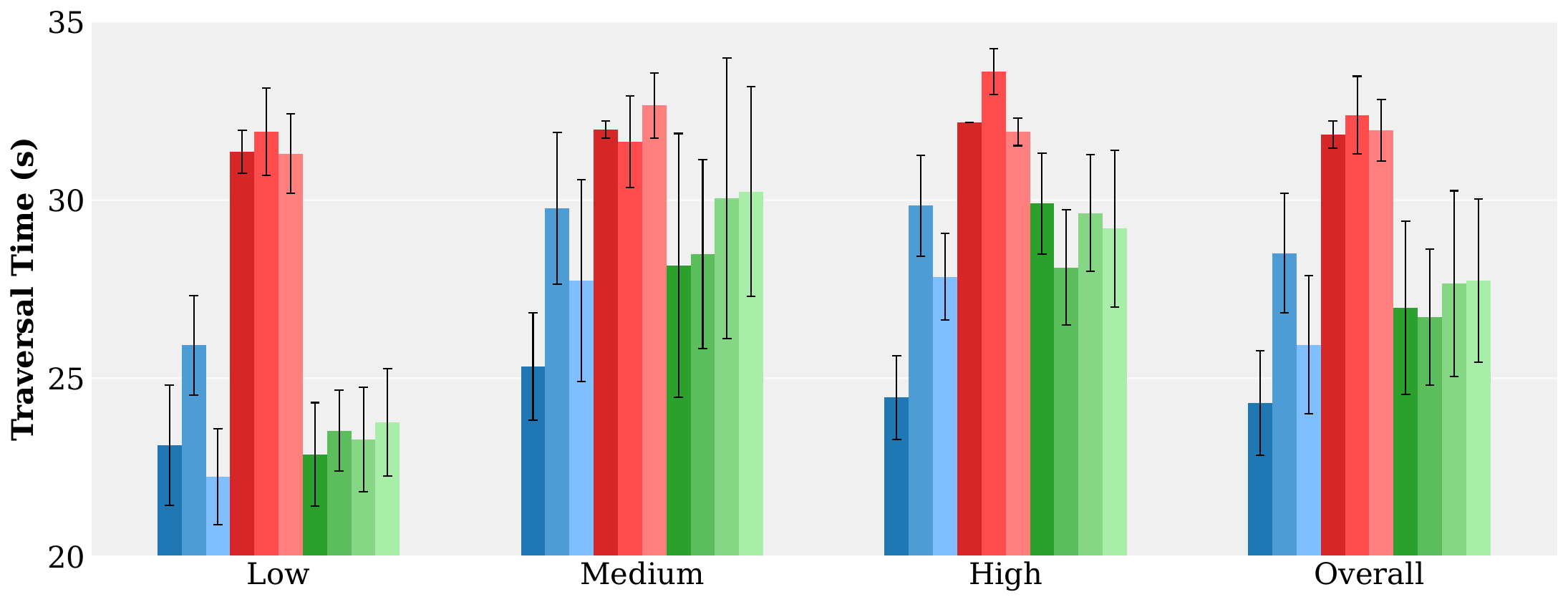}
    \includegraphics[width=0.49\textwidth]{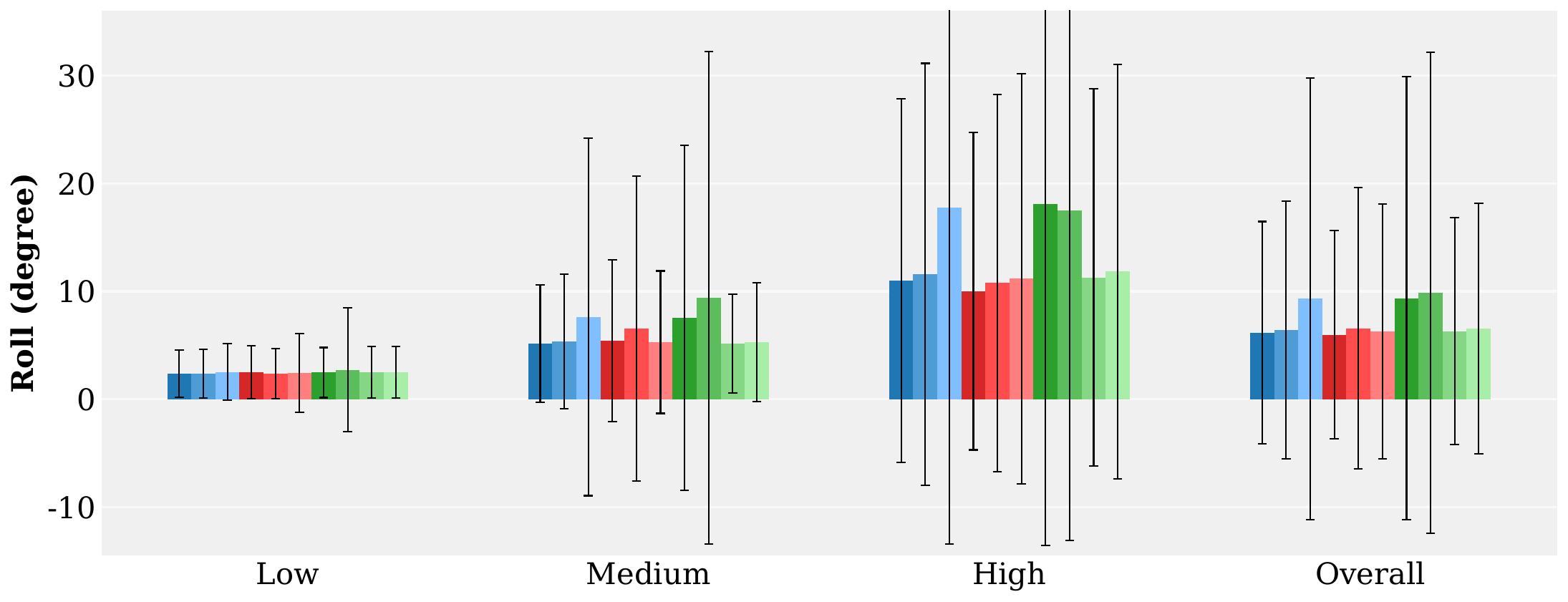}
    \includegraphics[width=0.49\textwidth]{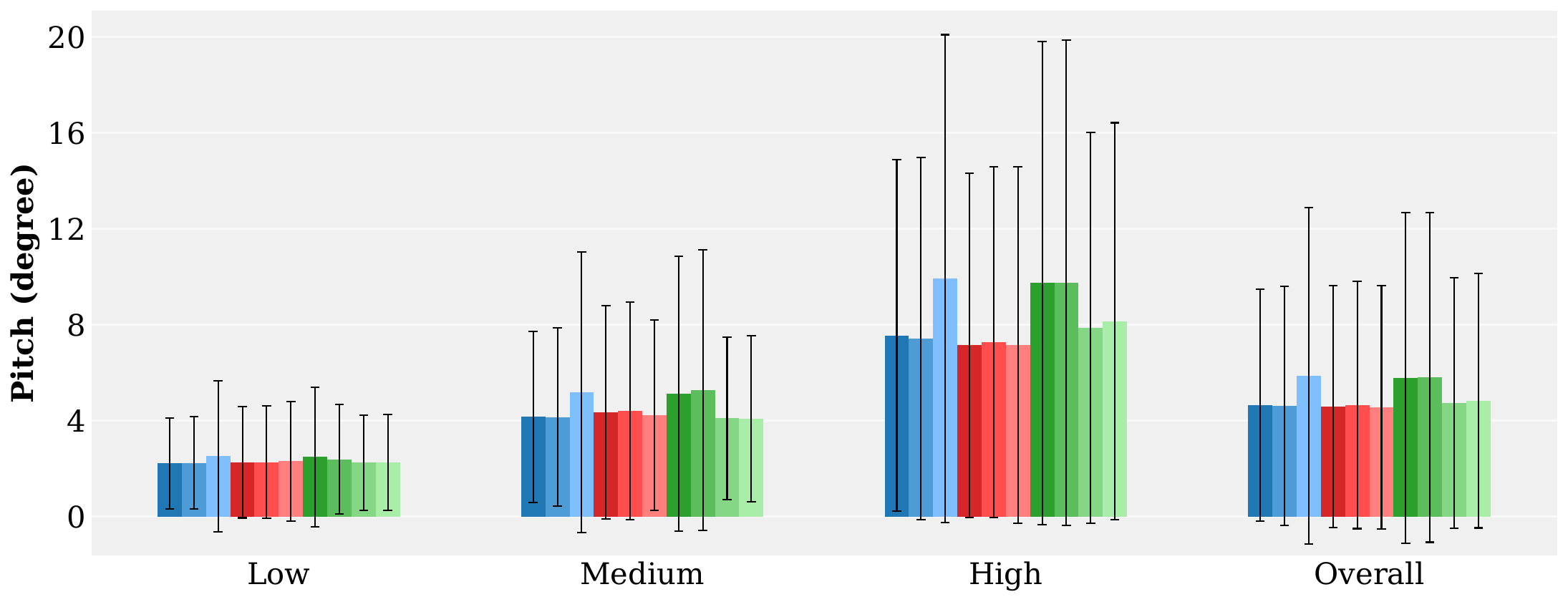}
    \includegraphics[width=0.49\textwidth]{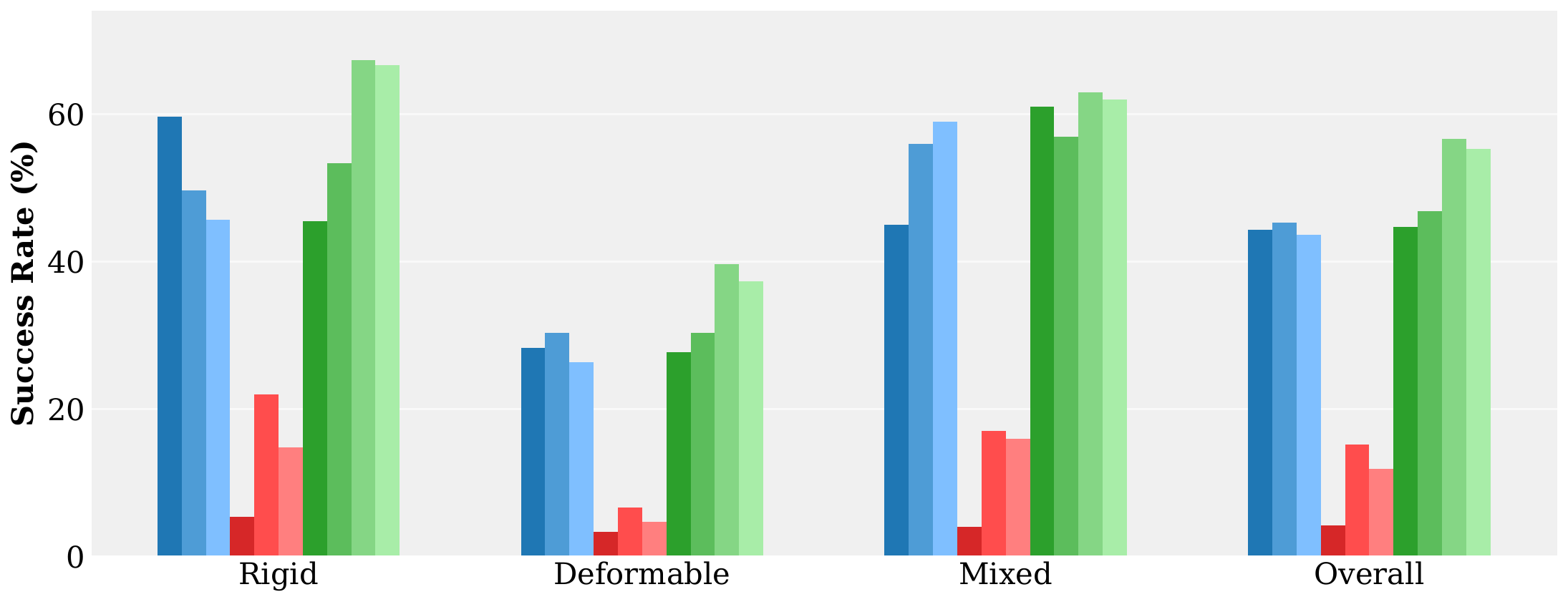}
    \includegraphics[width=0.49\textwidth]{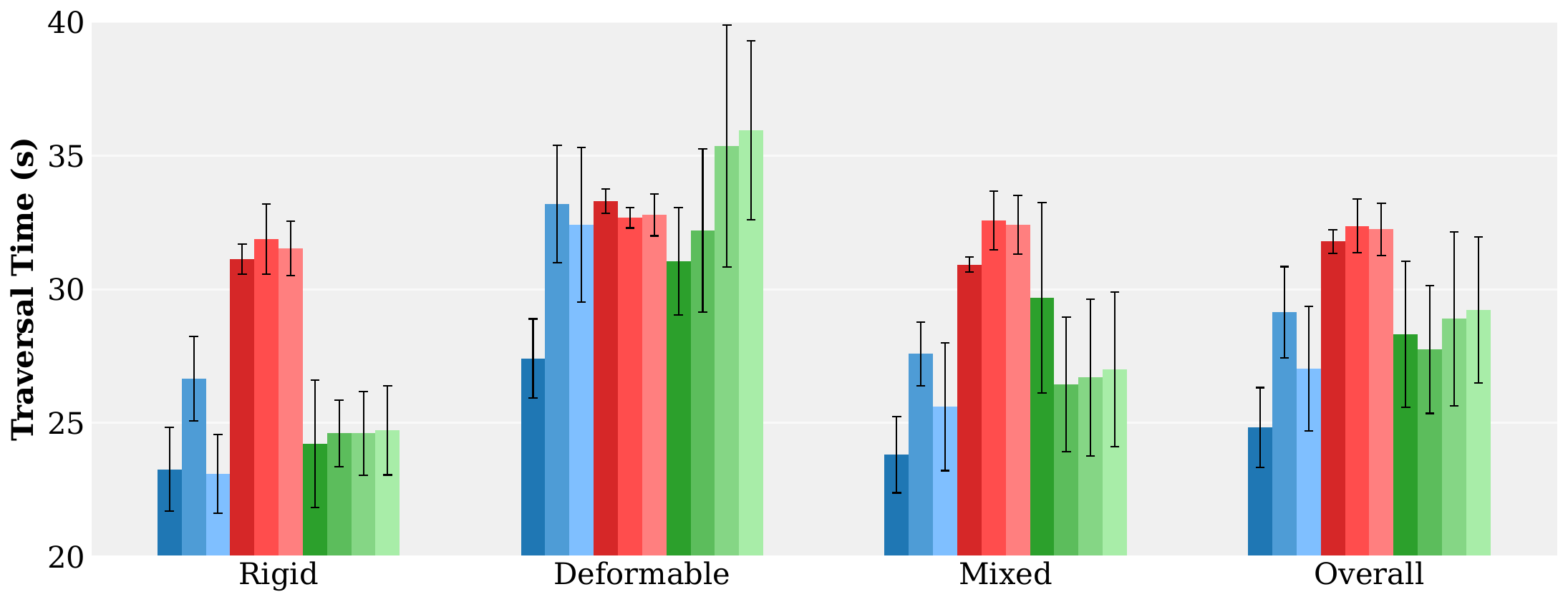}
    \includegraphics[width=0.49\textwidth]{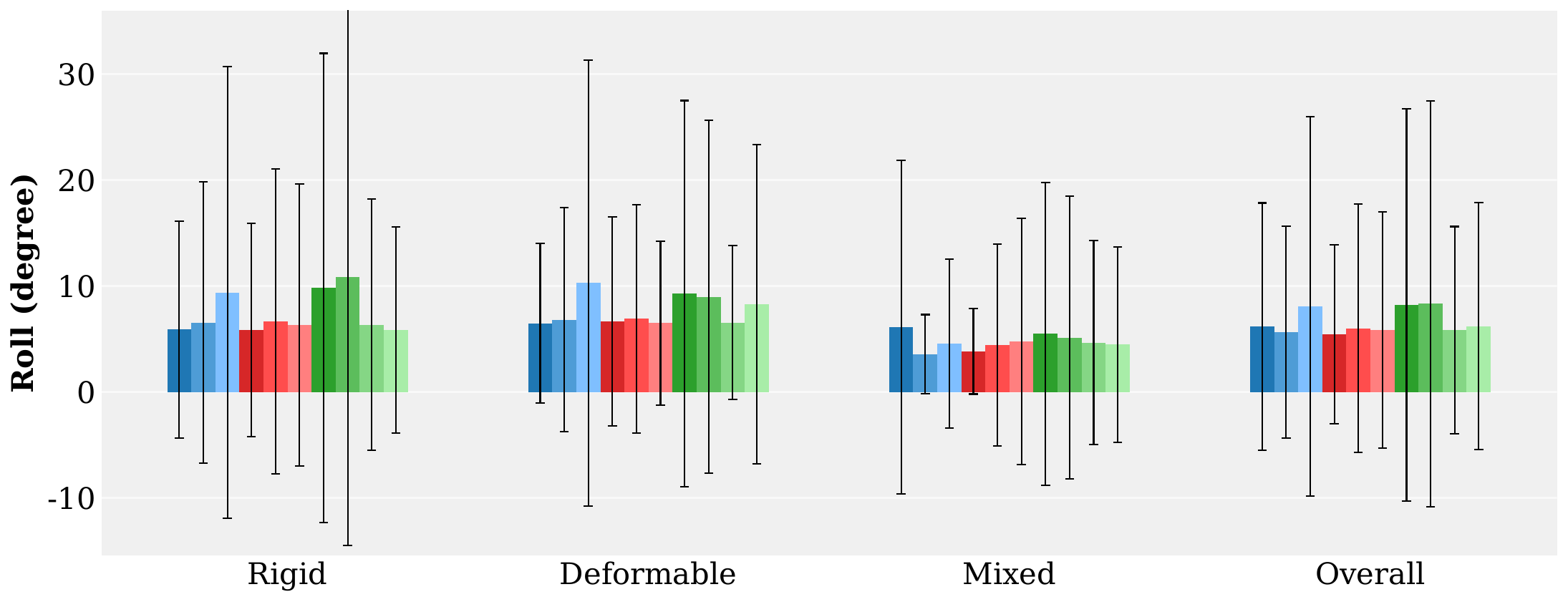}
    \includegraphics[width=0.49\textwidth]{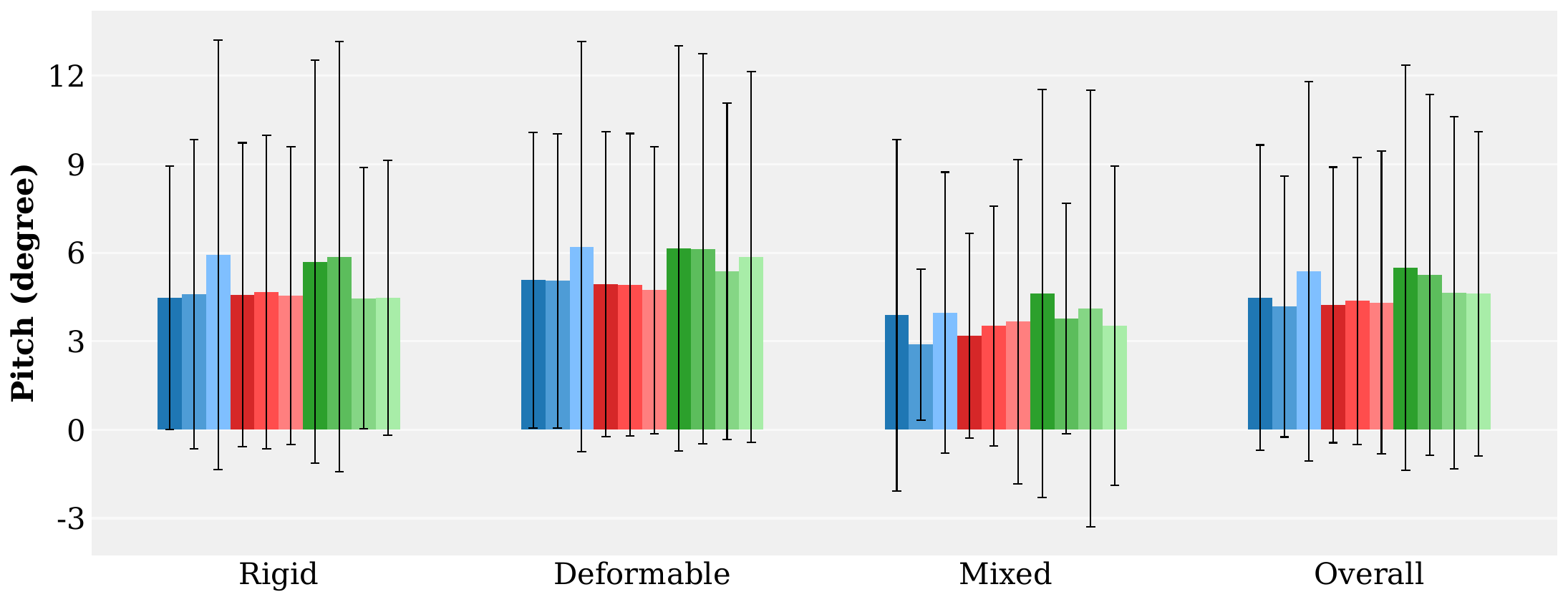}
    \includegraphics[width=0.49\textwidth]{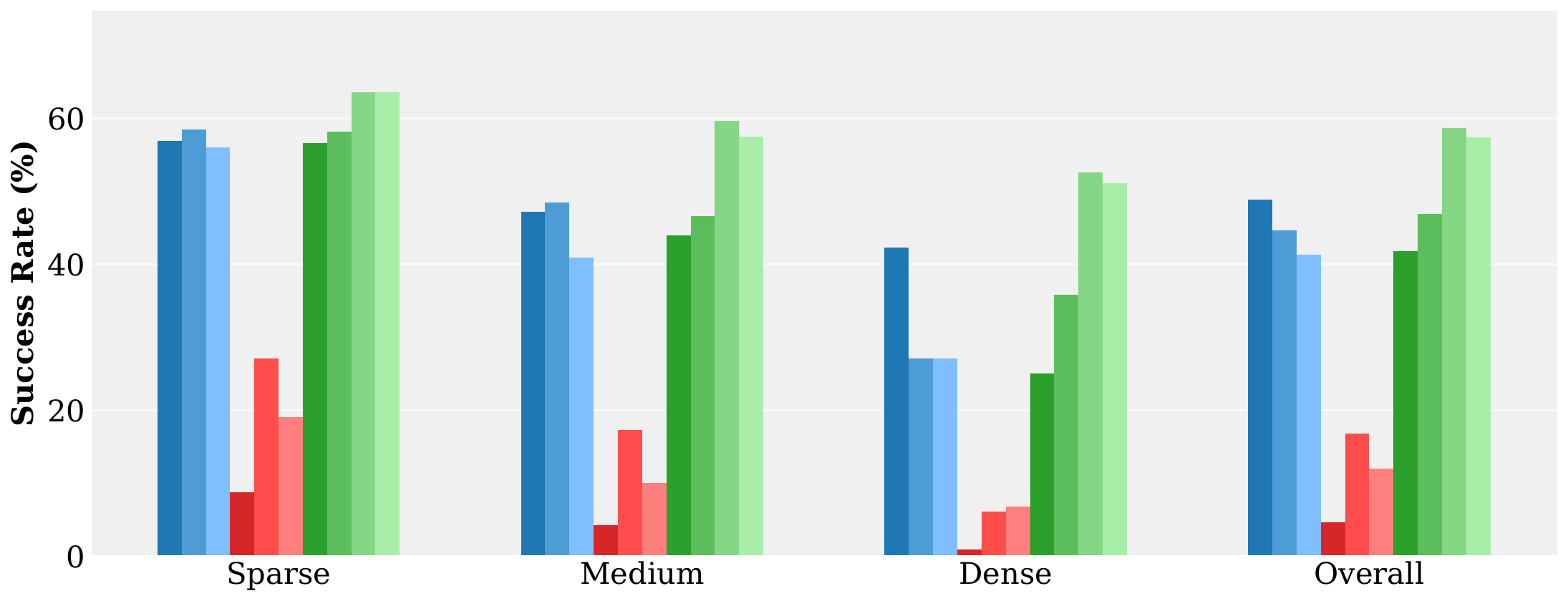}
    \includegraphics[width=0.49\textwidth]{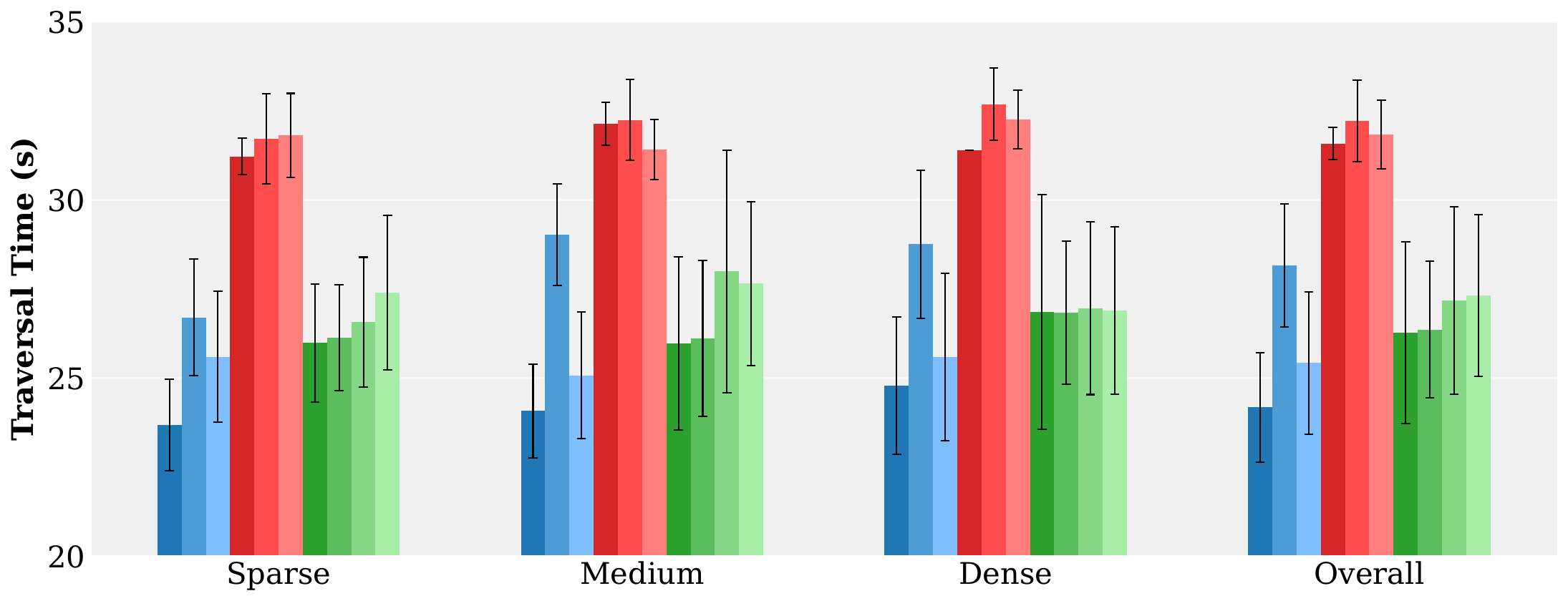}
    \includegraphics[width=0.49\textwidth]{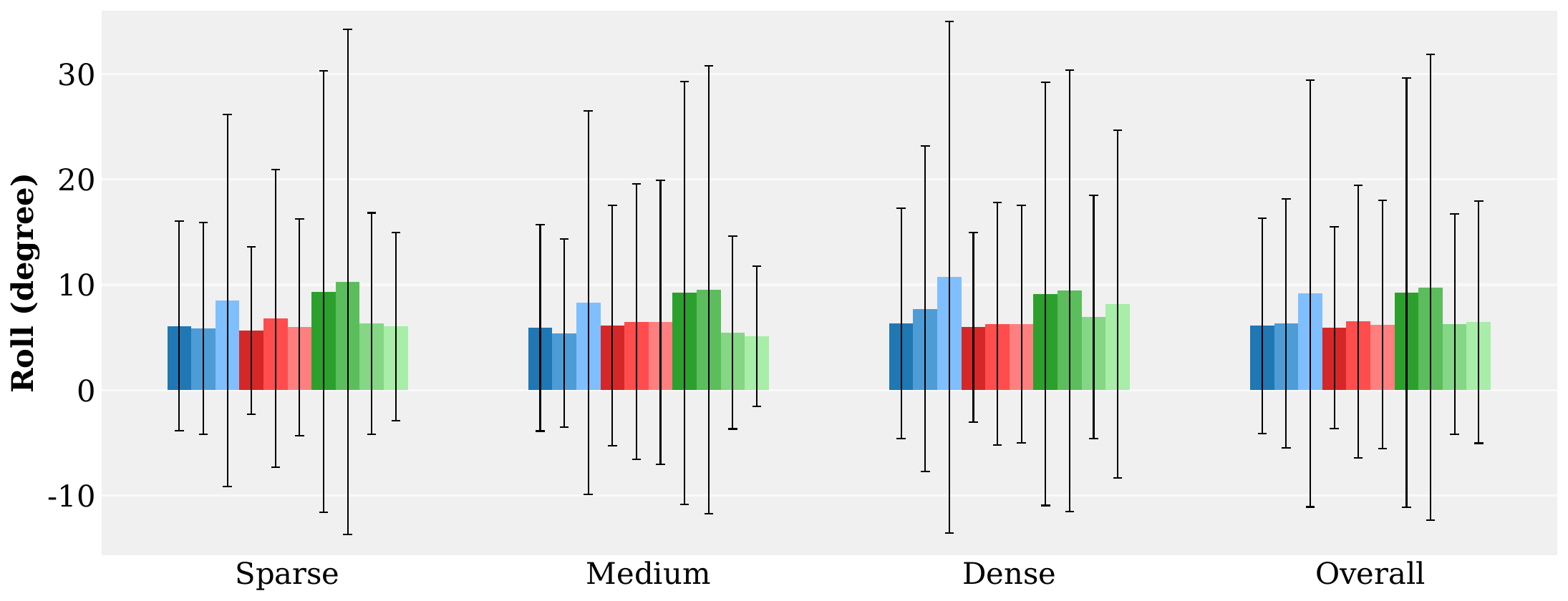}
    \includegraphics[width=0.49\textwidth]{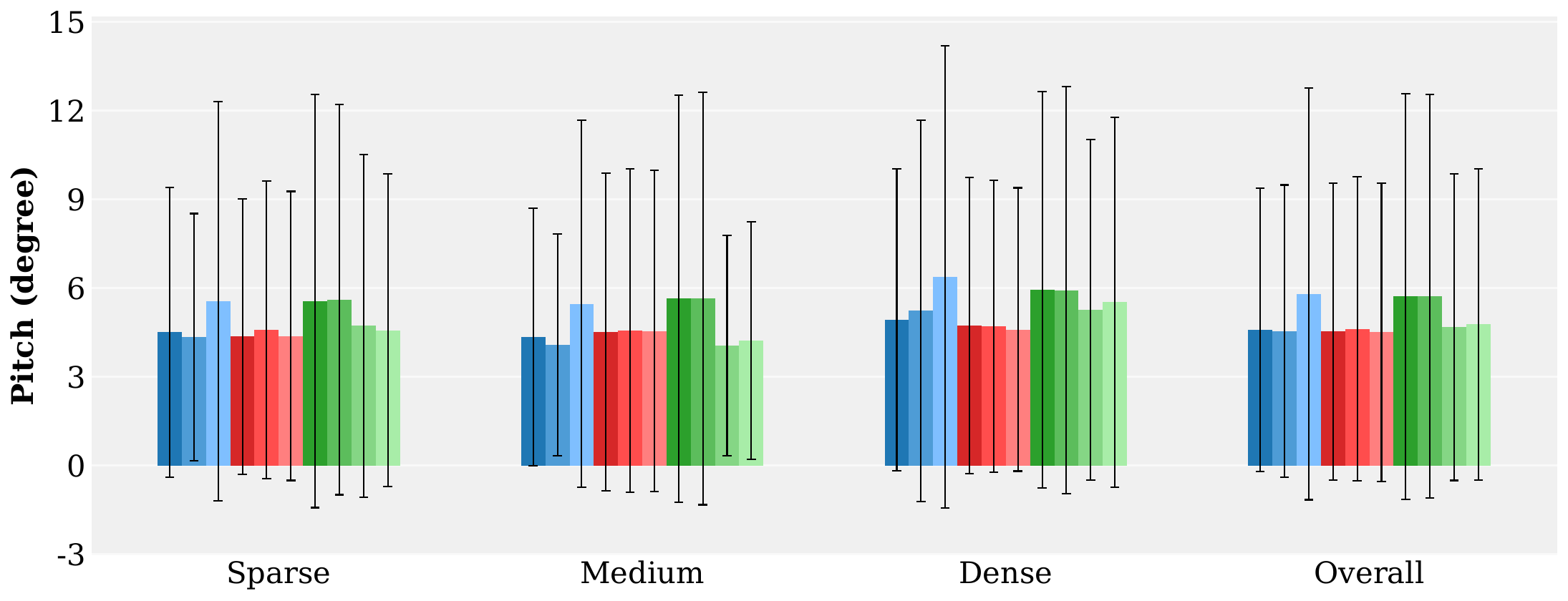}
    \includegraphics[width=1\textwidth]{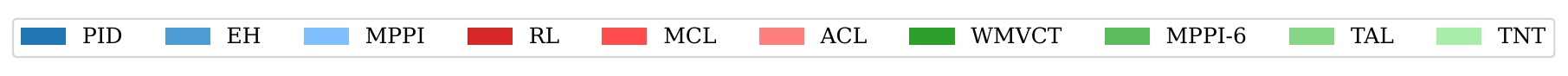}
    \caption{Success Rate, Traversal Time, Roll, and Pitch with respect to Elevation Level (top), Terrain Semantics (middle), and Obstacle Density (bottom) of Ten Off-Road Mobility Systems on 1000 Navigation Tasks.}
    \label{fig::eval}
\end{figure*}

We evaluate ten off-road mobility systems using Verti-Bench, ranging from purely classical, end-to-end learning, and hybrid systems. We present and discuss our evaluation results and point out future research directions. 

\subsection{Off-Road Mobility Systems for Evaluation}
The three classical off-road mobility systems include
\begin{itemize}
    \item PID: A controller that takes a local goal 10 m away from the robot on the global path and minimizes the error angle between the desired and vehicle heading by regulating the steering and maintaining a 3 m/s speed;
    \item Elevation Heuristics (EH): A controller that splits the elevation map in front of the current robot pose to five regions and drives toward the region with the most similar mean to the current terrain patch and lowest variance; 
    \item MPPI: An MPPI-based~\cite{williams2017model} planner that uses a 2D bicycle model for trajectory rollout and obstacle avoidance. 
\end{itemize}

The three systems based on end-to-end learning include
\begin{itemize}
    \item RL: A RL policy learned from trial and error~\cite{xu2024reinforcement}; 
    \item MCL: A RL policy learned from a manually designed curriculum~\cite{xu2024reinforcement};
    \item ACL: A RL policy learned using Automatic Curriculum Learning~\cite{xu2024verti}. 
\end{itemize}

The four hybrid (classical and learning) systems include

\begin{itemize}
    \item WMVCT: A planner based on a decomposed 6-DoF kinodynamic model (bicycle model for x, y, and yaw, elevation map for z, and neural network prediction for roll and pitch)~\cite{datar2024learning};
    \item MPPI-6: An MPPI-based planner with a learned full 6-DoF kinodynamic model for trajectory rollout~\cite{lee2023learning};
    \item TAL: An MPPI-based planner with a 6-DoF kinodynamic model that learns to attend to specific terrain patches~\cite{datar2024terrain};
    \item TNT: An MPPI-based planner that samples based on traversability and then unfolds 6-DoF kinodynamics~\cite{pan2024traverse}.
\end{itemize}
We reach out to the authors of the original papers for their implementations of their mobility systems. Considering Verti-Bench's focus on off-road mobility evaluation, we make minimal modifications to their implementations to interface with Verti-Bench so that their mobility systems are no longer dependent on any perception system. For example, visual odometry inputs are replaced with ground truth vehicle states from Verti-Bench; Real-world elevation mapping systems are skipped by directly providing their systems with ground truth Verti-Bench elevation maps; All learning systems and components in the evaluation are not trained in Verti-Bench.

\subsection{Evaluation Results and Discussions}

The evaluation results are shown in Fig.~\ref{fig::eval}, including percentage of Succuss Rate and mean and variance of Traversal Time, Roll, and Pitch with respect to three types of elevation level, terrain semantics, and obstacle density for all systems and tasks. We also present an analysis of failure cases shown in Table \ref{tab::failure}. All 1000 Verti-Bench navigation tasks have been used to evaluate each mobility system (no task has been used for training). A complete evaluation of each system requires approximately 10 hours. All task configurations have been documented in YAML files, which can be used by external research teams to replicate and expand these evaluations. Additionally, we provide our terrain generation pipeline, allowing researchers to extend or customize environmental parameters according to their specific research objectives and experiment requirements.

\begin{table}
\centering
\caption{Failure Case Analysis of 1000 tasks with Ten Mobility Systems}
\setlength{\tabcolsep}{1pt}
\renewcommand{\arraystretch}{1.3}
\label{tab::failure}
\begin{tabular}{ccccccccccc}
\toprule[1pt]
 & PID & EH & MPPI & RL & MCL & ACL & WMVCT & MPPI-6 & TAL & TNT \\
\midrule
Rollover (\%) & 14.7 & 16.3 & 17.9 & 14.3 & 16.9 & 15.4 & 18.9 & 18.0 & 17.0 & 17.5 \\
Stuck (\%) & 36.5 & 39.4 & 40.9 & 75.7 & 41.3 & 48.0 & 39.6 & 35.0 & 24.6 & 25.1 \\
Success (\%) & 48.8 & 44.3 & 41.2 & 10.0 & 41.8 & 36.6 & 41.5 & 47.0 & 58.4 & 57.4 \\
\bottomrule
\end{tabular}
\end{table}

In general, navigation performance significantly declines with increasing elevation levels, deformable surfaces, and obstacle densities, including reduced Success Rate and increased Traversal Time, Roll, and Pitch. In addition to mean, the variance of Roll and Pitch also drastically increases, indicating much less stable vehicle chassis when traversing high elevation, deformable, and obstacle-dense environments. Among the three categories, end-to-end learned mobility systems achieve the worst performance, while hybrid systems outperform the other two in general. While ACL is expected to outperform MCL, the results suggest otherwise. Such results indicate that end-to-end learning methods, trained from other sources, still have much room for improvement in terms of generalization in Verti-Bench. 
Considering failure cases, the systems fail more frequently due to getting stuck than rolling over. While the failure rates due to rollover are relatively consistent across all systems, failure rates due to getting stuck differ significantly, with RL getting stuck 75.7\% of time, compared to TAL and TNT with the lowest getting-stuck rates (24.6\% and 25.1\% respectively).

In terms of elevation, for hybrid systems, TAL and TNT are the two top performing planners among all systems overall, achieving the highest Success Rate in all cases and lowest Roll and Pitch angle in most cases. WMVCT and MPPI-6 achieve good Success Rate in high elevation environments, but with large Roll and Pitch. Classical planners perform in between their end-to-end and hybrid counterparts. PID, due to its simplicity and robustness, performs very well in low elevation environments, with EH catching up on Success Rate when facing higher elevation.  MPPI does not perform well in most cases and only outperforms PID in terms of Success Rate in high elevation environments. 
Notice that Traversal Time is only averaged over successful trials and thus only indicates how fast a mobility system is given navigation success. 

For terrain semantics, all systems perform best on rigid terrain, with TAL and TNT achieving highest Success Rates above 60\%. Performance drops significantly on deformable surfaces, with even the highest only achieving around 40\% Success Rates while end-to-end systems struggle below 10\%. Mixed terrain results fall between rigid and deformable terrain. Deformable surfaces also show increased roll and pitch variance across all systems, indicating less stable navigation and highlighting the challenge of modeling vehicle dynamics on unpredictable surfaces.

As obstacle density increases, Success Rate declines and Traversal Time increases across all systems, with the performance gap between hybrid and other systems widening in dense environments. Obstacle density does not directly affect vehicle stability, showing similar Roll and Pitch mean and variance.

Our evaluation results indicate the potential of hybrid mobility systems to tackle vertically challenging terrain by combining the best of both worlds of classical and learning approaches. The overall success of TAL and TNT indicates the importance of an accurate 6-DoF kinodynamic model enabled by sophisticated learning techniques in conjunction with a sampling-based motion planner. MPPI-6, with a 6-DoF kinodynamic model based on a simplistic neural network, underperforms TAL and TNT, while the inaccuracies introduced by WMVCT's efficient 6-DoF decomposition lead to the worst mobility performance among hybrid systems. The degraded performance of all hybrid systems when facing high elevation, deformable surfaces, and dense obstacles motivates further research, potentially to both increase the kinodynamic modeling accuracy and improve the sampling-based motion planner. On the other hand, it is surprising to see the superior performance of the simple PID planner in low elevation environments compared to the more sophisticated EH, whose advantage only starts to slightly emerge in high elevation environments. This observation reveals a tradeoff between system complexity and performance when facing simple environments. One potential future research direction is to develop off-road mobility systems composed of multiple planners with different complexities and specialties to fit different environments~\cite{choudhury2015planner}. Lastly, research of end-to-end learning approaches, despite their recent success in relatively benign indoor or on-road enviornments, still needs to focus on robustly generalizing to out-of-distribution scenarios, which are very common to encounter in off-road environments.

\section{Real-World Validation}
\label{sec::validation}

\begin{figure}[ht]
    \centering
    \includegraphics[width=\columnwidth]{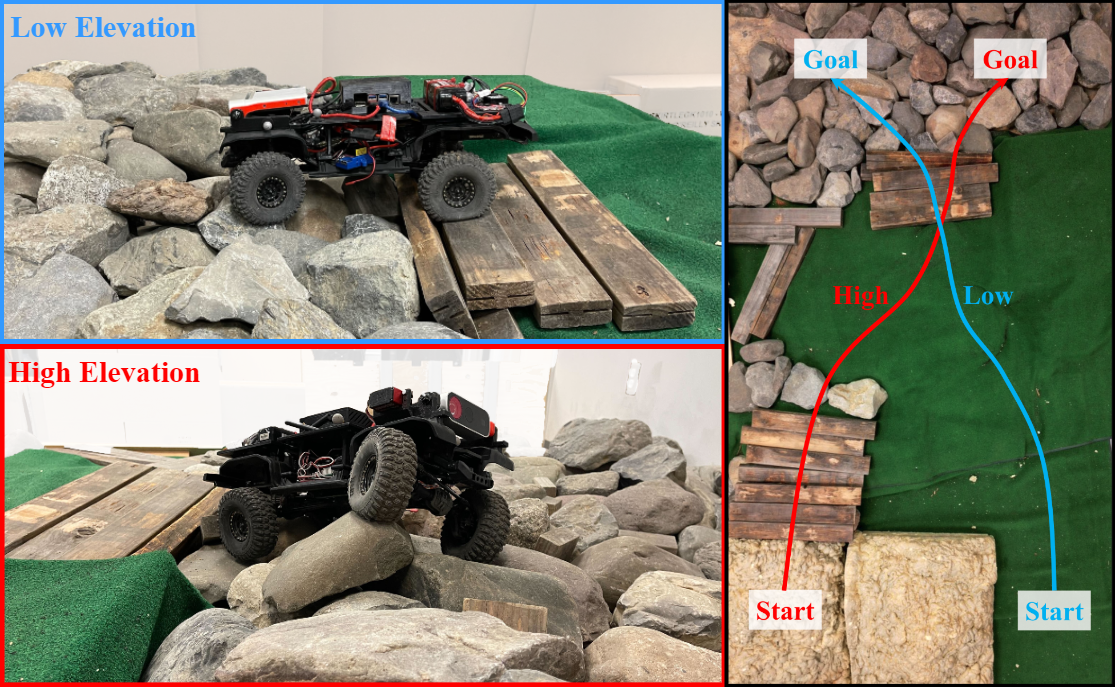}
    \caption{Physical Off-Road Testbed Similar to Verti-Bench.}
    \label{fig::real}
\end{figure}

To validate the Verti-Bench evaluation results, we deploy one representative mobility system from each of the three classes, i.e., PID for classical, ACL for end-to-end, and TNT for hybrid, on a physical 1/10th scale open-source Verti-4-Wheeler robot~\cite{datar2024toward} on an off-road mobility testbed constructed by rocks, foam, grass, and wood, presenting different geometry, semantics, and obstacle features (Fig.~\ref{fig::real}). The testbed is constructed in two different configurations to include low and high elevation in order to validate the cross-elevation evaluation results from Verti-Bench.  

Table \ref{tab::results} shows the physical validation results of the three mobility systems on both low and high elevation testbeds. In general, the trend of the physical experiment results matches with that of Verti-Bench evaluation results: On the low elevation testbed, both PID and TNT are able to finish all five trials, while ACL still suffers from poor generalization. PID is still the fastest due to a lack of consideration of elevation. The roll and pitch angles are small and stable for all systems due to the lower and smoother terrain elevation; 
On the high elevation testbed, the difference between TNT and PID starts to emerge. TNT succeeds all five trials, while PID fails two. TNT and PID also exhibit smaller roll and pitch angle respectively. PID is still the fastest. ACL, similar to all previous cases, fails three trials and experiences the largest roll and pitch angles. Notice that due to the difficulty in conducting physical experiments, we only limit to physically evaluating three systems and five trials each, totaling 30 trials. This contrast against our 10000 trials (ten systems, 1000 trials each) in Verti-Bench, which can achieve much better statistical significance, further suggests the utility of Verti-Bench to evaluate off-road mobility. 

\begin{table}
\caption{\textbf{Physical Validation of PID, ACL, and TNT:} Success Rate, Traversal Time, Roll, and Pitch.}
\centering
\resizebox{\columnwidth}{!}{%
\small
\setlength{\tabcolsep}{4pt}
\begin{tabular}{ccccccccc}
\toprule
Low Elevation                    & PID & ACL & TNT \\
\midrule
{Success Rate $\uparrow$}      & {\textbf{5/5}} & {3/5} & {\textbf{5/5}}\\
{Traversal Time $\downarrow$}      & {\textbf{6.64s$\pm$1.09s}} & {7.03s}$\pm$0.47s & {8.90s}$\pm$0.76s\\
{Roll $\downarrow$}      & {6.45}\textdegree$\pm$5.20\textdegree & {6.20\textdegree$\pm$4.13\textdegree} & \textbf{6.02\textdegree$\pm$4.92\textdegree}\\
{Pitch $\downarrow$}      & {5.45\textdegree$\pm$3.37}\textdegree & {6.34}\textdegree$\pm$3.32\textdegree & \textbf{{4.74\textdegree$\pm$3.23\textdegree}}\\
\midrule
High Elevation                   & PID & ACL & TNT  \\
\midrule
{Success Rate $\uparrow$}      & {3/5} & {2/5} & {\textbf{5/5}} \\
{Traversal Time $\downarrow$}     & \textbf{{11.00s}}$\pm$1.00s  & {16.00s}$\pm$0.72s & {17.50s$\pm$1.95s} \\
{Roll $\downarrow$}   & {10.14\textdegree$\pm$8.96\textdegree}   & {13.30\textdegree$\pm$12.72\textdegree}  & \textbf{{7.12}\textdegree$\pm$6.65\textdegree}\\
{Pitch $\downarrow$}   & \textbf{{7.61\textdegree$\pm$5.46\textdegree}}  & {9.78}\textdegree$\pm$6.76\textdegree  & {9.26\textdegree$\pm$8.41}\textdegree \\
\bottomrule
\end{tabular}%
}
\label{tab::results}
\end{table}

\section{Limitations}
\label{sec::limitations}
Despite being a general and scalable benchmark, Verti-Bench still has a few limitations. 
Due to the high requirement to compute high-fidelity physics and a lack of GPU acceleration, Verti-Bench can only achieve near-real-time simulation speed, with real time factor ranging between 0.4 and 1.5 (faster and slower than real time respectively) depending on simulation complexity. Integrating with GPU accelerators to increase benchmarking efficiency is an important next step. 
Verti-Bench aims to evaluate off-road mobility and assumes ground truth perception is available to the mobility system. However, such an assumption does not hold in the real world. Future work will add realistic perception noises and test the robustness of mobility systems when facing imperfect vehicle state estimation, elevation and semantics mapping, and obstacle detection. Another direction of expanding the current Verti-Bench is to create more complex real-world counterparts than the current small-scale physical testbed so that the sim-to-real gap can be more extensively studied to further validate the efficacy and improve the fidelity of Verti-Bench evaluation. 
\section{Conclusions}
\label{sec::conclusions}

We present Verti-Bench, a general and scalable off-road mobility benchmark for vertically challenging terrain. Based on Chrono, a high-fidelity multi-physics dynamics engine, Verti-Bench includes 100 off-road enviornments and 1000 navigation tasks with millions of off-road terrain features covering geometry, semantics, and obstacles. Verti-Bench can also scale to off-road vehicles of different sizes, weights, chassis, suspensions, and steering mechanisms. Standardized metrics and datasets are provided to quantify off-road mobility performance and facilitate data-driven mobility. Ten off-road mobility systems are evaluated in Verti-Bench, whose results are further validated on a physical testbed. We also point out future research directions to improve off-road mobility. 

\section*{Acknowledgments}
This work has taken place in the RobotiXX Laboratory at George Mason University. RobotiXX research is supported by National Science Foundation (NSF, 2350352), Army Research Office (ARO, W911NF2320004, W911NF2420027, W911NF2520011), Air Force Research Laboratory (AFRL), US Air Forces Central (AFCENT), Google DeepMind (GDM), Clearpath Robotics, Raytheon Technologies (RTX), Tangenta, Mason Innovation Exchange (MIX), and Walmart.

\bibliographystyle{IEEEtran}
\bibliography{references}


\end{document}